\documentclass{article}

\usepackage{PRIMEarxiv}
\usepackage{cite}

 \usepackage[pdftex]{graphicx}

\usepackage{amsmath}

\usepackage[caption=false,font=footnotesize]{subfig}
\usepackage{tabularx}

\usepackage{booktabs, caption}
\usepackage{svg}
\usepackage{soul}
\newcommand{\sectionname}{Section}
\renewcommand{\figurename}{Figure}
\usepackage{enumitem}
\newcommand{\tr}{\toprule}
\newcommand{\br}{\bottomrule}
\newcommand{\mr}{\midrule}
\usepackage[utf8]{inputenc} 
\usepackage[T1]{fontenc}    
\usepackage{hyperref}       
\usepackage{url}            
\usepackage{amsfonts}       
\usepackage{nicefrac}       
\usepackage{microtype}      
\usepackage{lipsum}
\usepackage{fancyhdr}       
\usepackage{graphicx}       
\graphicspath{{media/}}     

\title{Simple and complex spiking neurons: perspectives and analysis in a simple STDP scenario}

\author{Davide Liberato~Manna, Alex~Vicente Sola,\\
{\bf Paul~Kirkland, Gaetano~Di Caterina} \\
Neuromorphic Sensor Signal Processing Lab\\
Centre for Image and Signal Processing\\ 
Electrical and Electronic Engineering\\
University of Strathclyde\\
Glasgow, United Kingdom. \\
\texttt{email: davide.manna@strath.ac.uk}
\And
 Trevor~Bihl\\
 Air Force Research Laboratory\\
 Wright Patterson AFB, OH}

\begin{document}

\maketitle

\begin{abstract}
Spiking neural networks (SNNs) are largely inspired by biology and neuroscience and leverage ideas and theories to create fast and efficient learning systems. Spiking neuron models are adopted as core processing units in neuromorphic systems because they enable event-based processing. Among many neuron models, the integrate-and-fire (I\&F) models are often adopted, with the simple Leaky I\&F (LIF) being the most used. The reason for adopting such models is their efficiency and/or biological plausibility.
Nevertheless, rigorous justification for adopting LIF over other neuron models for use in artificial learning systems has not yet been studied. This work considers various neuron models in the literature and then selects computational neuron models that are single-variable, efficient, and display different types of complexities. From this selection, we make a comparative study of three simple I\&F neuron models, namely the LIF, the Quadratic I\&F (QIF)  and the Exponential I\&F (EIF), to understand whether the use of more complex models increases the performance of the system and whether the choice of a neuron model can be directed by the task to be completed.
Neuron models are tested within an SNN trained with Spike-Timing Dependent Plasticity (STDP) on a classification task on the N-MNIST and DVS Gestures datasets. Experimental results reveal that more complex neurons manifest the same ability as simpler ones to achieve high levels of accuracy on a simple dataset (N-MNIST), albeit requiring comparably more hyper-parameter tuning. However, when the data possess richer Spatio-temporal features, the QIF and EIF neuron models steadily achieve better results. This suggests that accurately selecting the model based on the richness of the feature spectrum of the data could improve the whole system's performance. Finally, the code implementing the spiking neurons in the SpykeTorch framework is made publicly available.

\end{abstract}

\noindent{\it Spiking Neural Networks, STDP, Unsupervised Learning, Spiking Neurons, Temporal Features\/}


%

\section{Introduction}\label{sec:introduction}

%
%
%
%
As technology in the Neuromorphic (NM) computing field keeps on advancing, so are the software methodologies and algorithms that can leverage the low-power, low-latency and event-driven properties that characterize NM \cite{Schuman2017ASurveyofNeuromorphicComputingandNeuralNetworksinHardware}. Often, inspired by the success of conventional deep learning, this results in the development of Spiking Neural Networks (SNNs).
When it comes to designing an SNN learning system for some machine learning task, researchers are faced with many decisions to make. Among these comes the choice of a particular neuron model. This specific aspect of the development of an SNN is a very sensitive one as spiking neurons are the core processing units of an NM system. To draw a parallel with the conventional Deep Learning (DL) research, spiking neurons can be thought of as being activation functions (such as the ReLU, ATAN etc), but holding an internal state. The dynamics of this state through time are governed by the differential equations that constitute the spiking neuron model.

Different neuron models exhibit different state dynamics. From a neuroscience point of view, these differences are very clear \cite{Gerstner_2009}. Some models are able to capture certain intrinsic behaviours of neurons, e.g. they can burst, chatter or fast-spike, while others cannot. Some models are also better at approximating subthreshold dynamics, thus possibly being more accurate representations of real neurons.
However, it is still unclear how this ability translates into applicability in SNNs. There is in fact no definite answer onto whether certain types of neural dynamics can be beneficial to particular SNN applications, nor any common knowledge on the criteria that should drive the choice of such neurons in relation to such dynamics.

Within some specific contexts, the choice is constrained by the available hardware. As a matter of fact, several neuromorphic chips allow to only adopt the specific neuron model that the chip is able to emulate. BrainScaleS \cite{Schemmel2020AcceleratedAnalogNeuromorphicComputing} for instance allows to only adopt the Adaptive Exponential Integrate-and-Fire neuron model; NeuroGrid \cite{NeurogridAMixedAnalogDigitalMultichipSystem} allows only an Adaptive Quadratic Integrate-and-Fire model ; meanwhile, Loihi \cite{LohiANeuromorphicManycoreProcessorWithOnChipLearning}, SyNAPSE \cite{ADigitalNeurosynapticCoreUsingEmbeddedCrossbarMemory, CognitiveComputingBuildingBlockAVersatileAndEfficientDigitalNeuronModel} and TrueNorth \cite{Akopyan2015TruenorthDesignAndToolFlowOfA} allow only  Leaky Integrate-and-Fire (LIF) based models.
Until recently, the only chip allowing the implementation of any type of neuron model was SpiNNaker \cite{Furber2014TheSpinnakerProject}. However, the recently released Loihi 2 adds to the list of chips with programmable neuron models \cite{orchard2021EfficientNeuromorphicSignalProcessingWithLoihi2}, hence highlighting the importance of an accurate investigation on this matter.

It is thus interesting to look at what are the most suitable neurons models for SNN development. This can help to understand if the dynamics of the neurons relate, in some way, to the dynamics of the spatio-temporal features of the data.
\\\\
Simple LIF neurons are the de-facto standard choice when it comes to SNN design \cite{VicenteSola2021KeystoAccurateFeatureExtractionUsingResidualSpikingNeuralNetworks, Aamir2018AnAcceleratedLIFNeuronalNetworkArrayforaLargeScaleMixedSignalNeuromorphicArchitecture, Diamond2016ComparingNeuromorphicSolutionsinActionImplementingaBioInspiredSolutiontoaBenchmarkClassificationTaskonThreeParallelComputingPlatforms, Friedl2016HumanInspiredNeuroroboticSystemforClassifyingSurfaceTexturesbyTouch, Hunsberger2015SpikingDeepNetworkswithLIFNeurons, Goeltz2021FastandEnergyEfficientNeuromorphicDeepLearningwithFirstSpikeTimes, Mozafari2018FirstSpikeBasedVisualCategorizationUsingRewardModulatedSTDP, Stromatias2015ScalableEnergyEfficientLowLatencyImplementationsofTrainedSpikingDeepBeliefNetworksonSpiNNaker, Mostafa2018SupervisedLearningBasedonTemporalCodinginSpikingNeuralNetworks, chaturvedi2012ImageSegmentationUsingLeakyIntegrateAndFireModelOfSpikingNeuralNetwork, jiang2021ASpikingNeuralNetworkWithSpikeTimingDependentPlasticityForSurfaceRoughnessAnalysis, PatinoSaucedo2020EventDrivenImplementationOfDeepSpikingConvolutionalNNForsupervisedClassificationUsingTheSpiNNaker}. When a rationale for this is provided, this choice is often attributed to the simplicity and, consequently,  to the efficiency of the LIF neuron model. Whatever the case, such reasons hardly account for the accuracy performance of the task at hand, neither for the temporal feature representation capability of the model. Other works rely on more complex neuron models \cite{Fardet2018UnderstandingtheGenerationofNetworkBurstsbyAdaptiveOscillatoryNeurons, Taherkhani2017AnArtificialNeuralNetworkBasedonIzhikevichNeuronModel, Chaturvedi2014ReviewofHandwrittenPatternRecognitionofDigitsandSpecialCharactersUsingFeedForwardNeuralNetworkandIzhikevichNeuralModel, Vazquez2011TrainingSpikingNeuralModelsUsingCuckooSearchAlgorithm}, attributing the choice to the biological plausibility or, once again, to efficiency. 
Neurons with different dynamics are present in areas of the brain with different functionalities \cite{Markram2004InterneuronsoftheNeocorticalInhibitorySystem,Connors1990IntrinsicFiringPatternsofDiverseNeocorticalNeurons, Dauth2017NeuronsDerivedFromDifferentBrainRegionsAreInherentlyDifferentInVitro}; however, the same does not apply to spiking neuron models in ML, where often the simplest neurons are used, leaving it unclear whether there are advantages or disadvantages relative to different types of NM data. 
\\\\
This work aims to answer the following research questions:
\begin{itemize}[itemsep=4pt]
    \item Does the chosen neuron model influence the performance of an SNN?
    \item Should the choice of the neuron model be related with the data it will have to process?
\end{itemize}

Specifically, we are interested in understanding whether neuron models with different and more complex neuronal dynamics display any advantages over simpler (LIF) ones in an unsupervised learning context. Furthermore, we investigate whether such differences might exist depending on the task set to the network; hence we perform experiments using two different datasets. To do this, we first develop a basic experiment, which represents a simple yet efficient way to start a comparison between neuron models. We select neuron models so that they scale up in terms of complexity and spiking patterns, and evaluate their performance within the same neural network architectures on the N-MNIST dataset \cite{Orchard2015ConvertingStaticImageDatasetstoSpikingNeuromorphicDatasetsUsingSaccades}. Then, we use the same neural network and train it on classification tasks taken from the DVS Gestures dataset \cite{Amir2017ALowPowerFullyEventBasedGestureRecognitionSystem}, which displays a richer distribution of events.
In order to perform the aforementioned experiments, we further contribute by enriching the SpykeTorch \cite{Mozafari2019SpykeTorch} framework with a new set of spiking neurons\footnotemark{}.
\footnotetext{Code available at https://www.github.com/daevem/SpykeTorch-Extended}
\\\\
The rest of this paper is organized as follows: in \sectionname\ 2 we provide some background information regarding the multitude of neuron models found in the literature and highlight some relevant related works; in \sectionname\ 3 we present our experimentation pipeline in detail, focusing on the datasets, neural network design and learning paradigms; \sectionname s 4 and 5 contain respectively the results obtained through our experiments and initiate an in-depth discussion on such results; \sectionname\ 6 concludes the paper.

\section{Background and Related Works}

Spiking neuron models were first born in the field of neuroscience and neurophysiology, where mathematical models were developed to reproduce what was found by recording the activity of real neurons \cite{Lapicque1907RecherchesQuantitativesSurLExcitationElectriqueDesNerfsTraiteeCommeUnePolarization, Hodgkin_1952, Gerstner2002SpikingNeuronModels}. Early spiking neural networks resulted from studies aimed at understanding biological dynamics \cite{gerstner1993ABiologicallyMotivatedAndAnalyticallySolubleModelOfCollectiveOscillationsInTheCortex},  or simulating areas of the brain and neuronal interactions \cite{van1996ChaosInNeuronalNetworksWithBalancedExcitatoryAndInhibitoryActivity, kawato1992AComputationalModelOfFourRegionsOfTheCerebellum, Zipser1993ASpikingNetworkModelOfShortTermActiveMemory}. The shift towards their use for computational tasks was gradual and was formalised afterwards \cite{maass1996LowerBoundsForTheComputationalPowerOfNetworksOfSpikngNeurons, Maass1997NetworksofSpikingNeuronstheThirdGenerationofNeuralNetworkModels}, with the focus generally being on the overall computational abilities of the network.
When it comes to developing SNNs for ML applications, spiking neurons can be thought of as stateful activation functions. This means they retain a state of their value (the membrane potential) reached through previous inputs. They are thresholding functions, therefore allowing to only forward information upon the reaching of a set threshold
. 

In conventional DL, activation functions have been extensively studied due to their importance in the propagation of the information. Nonlinear functions such as the sigmoid and Tanh were introduced to break the linearity of multilayer Perceptrons \cite{Goodfellow-et-al-2016, sharma2017activationFunctionsInNeuralNetworks}. Rectified Linear Units (ReLUs) substituted them to solve the vanishing gradient problem and allow deeper networks. Further variants \cite{Maas2013RectifierNonlineraitiesImproveNeuralNetwork, He2015DelvingDeepIntoRectifiersParametricRelu, Clevert2016FastAndAccurateDeepNetworkLearningByELU, Ramachandran2018SearchingForActivationFunctions} addressed other issues like the dying ReLU problem and helped to improve the performance of the networks \cite{pedamonti2018comparisonOfNonLinearActivationFunctionsForDeepNeuralNetworks, eger2019IsItTimeToSwishComparingDLActivationFunctions, Din2018ActivationFunctionsAndTheirCharacteristicsInDeepNeuralNetworks, zheng2020rethinkingTheRoleOfActivationFunctionsInDeepConv, Goyal2020ActivationFunctions, JAGTAP2020AdaptiveActivationFunctionsAccelerateConvergence}.
Instead, in the context of SNNs and spiking neurons, it is hard to find works in the literature relative to the differences in the use of different neuron models in SNNs for NM and DL applications. To the best of our knowledge, the only work considering the role of spiking neurons in learning from a DL point of view is the one by Traub et al. \cite{Sboev2018TotheRoleoftheChoiceoftheNeuronModelinSpikingNetworkLearningonBaseofSpikeTimingDependentPlasticity};  however, they focus on the qualitative properties of Spike-Timing-Dependent Plasticity (STDP) and the mean firing rate of the system after training. Furthermore, they consider a non-NM dataset and a different set of neurons.

Most of the works on spiking neurons concentrate on the neurobiological aspects they expose. One of the most influential works in this matter is the one by Izhikevich \cite{Izhikevich_2004}, which compared several models of spiking neurons (LIF, LIF with adaptation, LIF-or-burst, resonate-and-fire, QIF, Izhikevich's, FitzHugh-Nagumo, Hindmarsh-Rose, Morris-Lecar, Wilson, Hodgkin-Huxley), outlining their ability to reproduce observed neuronal behaviours and the cost (in terms of floating-point operations) of implementing such neurons in software applications. Similar work was conducted in  \cite{Long2010AReviewofBiologicallyPlausibleNeuronModelsforSpikingNeuralNetworks}, but focusing on a smaller subset of neurons (LIF, Izhikevich's, FitzHugh-Nagumo, Wilson, Hodgkin-Huxley) and analysing their numerical stability.
Although they closely study spiking neuron models, the two studies above concentrate on their computational costs and the intrinsic biological mechanics that each model can reproduce. However, they do not consider the effect of using spiking neurons with different dynamics in a DL system.
\\
A number of other works concentrate on the efficacy of neuron models in representing observed cortical neurons firing patterns. One example is given by  \cite{Barton2018TheApplicationPerspectiveofIzhikevichSpikingNeuralModeltheInitialExperimentalStudy}, where the authors make an exploratory analysis of how parameters influence Izhikevich's neurons in showing different spiking patters. Still regarding Izhikevich's neurons, Kumar et al. \cite{OptimalParameterEstimationOfTheIzhikevishSingleNeuronModel} estimate parameters that allow the neuron model to optimally reproduce a given spike train. Teeter et al. \cite{Teeter2018GeneralizedLeakyIntegrateandFireModelsClassifyMultipleNeuronTypes} use a generalized version of the LIF neuron model (GLIF) to understand whether more complex models allow to predict spike timing behaviors more closely; they conclude that this ability does not increase monotonically with the complexity, nor with the ability to reproduce sub-threshold dynamics. In  \cite{IntegrateAndFireModelsWithAdaptationAreGoodEnough} it is argued that integrate-and-fire (I\&F) neuron models are good enough estimators of input spike trains when coupled with an adaptation variable. This is both quantitatively and qualitatively shown by the authors and provides a good ground to the adoption of this family of neuron models.\\
Finally, neuron models in  \cite{Grzyb2009WhichModeltoUsefortheLiquidStateMachine} are compared within a retina-like neural network modelled with a liquid state machine (LSM). The performance of each neuron here is measured in terms of the "separation ability", i.e. the ability of the LSM to generate different responses to different stimuli. They find that all the models achieved reasonable separation ability, with the exception of Izhikevich's model.

\begin{figure}[!t]
\centering
\includegraphics[width=\columnwidth]{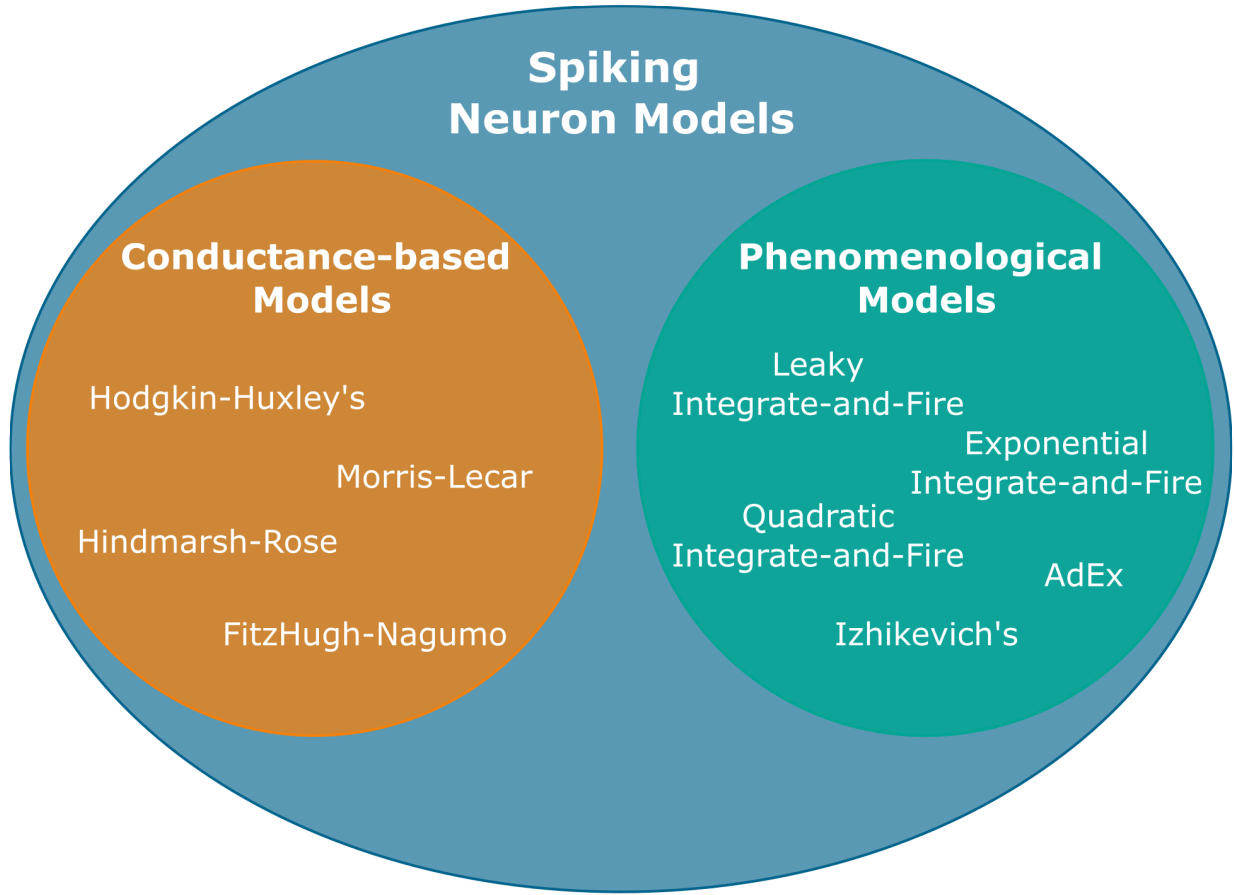}
\caption{Venn Diagram of Some Spiking Neurons.}
\label{fig:venn_neurons}
\end{figure}
\subsection{Neuron Models in the Literature}\label{sec:background_neurons}
The neuroscience literature is full of different neuron models, mostly describing the dynamics of the soma (core) of the cortical neurons. These can be roughly subdivided into two larger groups (see \figurename\ \ref{fig:venn_neurons}), the bio-physical or conductance-based models and the event-based or integrate-and-fire models.
\subsubsection{Conductance-based Models}
This class of neuron models is characterized by the fact that all the variables and parameters present in the model have a biophysical correspondence and are therefore measurable through experiments \cite{Izhikevich2010DynamicalSystemsinNeuroscience}. Among them, the Hodgkin-Huxley (HH) model is considered to be one of the most important in computational neuroscience and defines a system of 4 non-linear differential equation with four variables and a number of parameters. While further levels of complexity can be attained by including further variables in the model, this is not amenable to mathematical analysis. In fact, other simpler conductance-based models have been derived in the literature in order to ease the analysis, while still retaining biophysical plausibility. Some examples are the FitzHugh-Nagumo model \cite{FitzHugh1961ImpulsesandPhysiologicalStatesinTheoreticalModelsofNerveMembrane, Nagumo1962AnActivePulseTransmissionLineSimulatingNerveAxon}, the Hindmarsh-Rose \cite{Hindmarsh1984AModelofNeuronalBurstingUsingThreeCoupledFirstOrderDifferentialEquations} and the Morris-Lecar model \cite{Morris1981VoltageOscillationsintheBarnacleGiantMuscleFiber}. Nevertheless, they still remain rather complex for what concerns analysis and computation, therefore this family of neuron models is often used only when studying single-cell or small population dynamics \cite{Izhikevich_2004}.
\subsubsection{Phenomenological Models}
The family of Integrate-and-Fire or phenomenological neuron models comprises all those models that treat spikes as stereotypical events in time \cite{Gerstner_2009}. Therefore, each spike is completely described by the time at which it occurred, or was emitted. Integrate-and-fire models require at least two equations, one describing the dynamics of the membrane potential and the other one defining the action potential generation. Events are integrated over time and convey electrical charges that can cause excitation or inhibition of the membrane potential of the receiving neuron.
Differently from the conductance-based models, the phenomenological ones are more indicated for the development of neural networks \cite{Burkitt2006AReviewoftheIntegrateandFireNeuronModelI.HomogeneousSynapticInput, Gerstner2009HowGoodAreNeuronModels, Izhikevich_2004}, thanks to their overall lower complexity and the lower number of parameters, which enable easier fitting. \\\\
The simplest model, apart from the perfect integrator, is the LIF \cite{Lapicque1907RecherchesQuantitativesSurLExcitationElectriqueDesNerfsTraiteeCommeUnePolarization}. The dynamics of the membrane potential are here described by the following linear differential equation:
\begin{equation}
\label{LIF_equation}
    \tau_m\frac{du}{dt} = -(u(t)-u_{rest}) + R\cdot I
\end{equation}
where $\tau_m$ is the membrane time constant, $u(t)$ is the membrane potential as a function of time, $u_{rest}$ is the resting potential of the membrane, $R$ is a resistance and $I$ is the incoming current.

Although this model lacks the ability to describe most of the neuronal dynamics, it is the most common choice for the development of large scale neural networks, mostly because of its efficiency.

More complex I\&F models attempt to account for some non-linear dynamics of neurons as a function of the value of their membrane potential in a certain moment in time. Two examples are given by the Exponential Integrate-and-Fire (EIF) model \cite{FourcaudTrocme2003HowSpikeGenerationMechanismsDeterminetheNeuronalResponsetoFluctuatingInputs} and by the Quadratic Integrate-and-Fire neuron (QIF) or Theta neuron \cite{Ermentrout1986ParabolicBurstinginanExcitableSystemCoupledwithaSlowOscillation, Izhikevich2010DynamicalSystemsinNeuroscience}. 
As shown in (\ref{EIF_equation}), EIF model expands on the LIF model by including an exponential dependency on the current state of the membrane potential:
\begin{equation}
    \label{EIF_equation}
    \tau_m\frac{du}{dt} = -(u(t)-u_{rest}) + \Delta_T\exp{\left(\frac{u(t)-\Theta_{rh}}{\Delta_T}\right)} + R\cdot I,
\end{equation}
where $\Delta_T$ is a parameter determining the sharpness of the exponential curve and $\Theta_{rh}$ is the rheobase threshold. When $u > \Theta_{rh}$, the exponential term becomes prominent over the linear one, leading to an upswing of the curve that takes the membrane potential to infinity in finite time.

The QIF model, given by (\ref{QIF_equation}), employs a quadratic dependency from the membrane potential:
\begin{equation}
    \label{QIF_equation}
    \tau_m\frac{du}{dt} = a_0(u(t)-u_c)(u(t)-u_{rest}) + R\cdot I,
\end{equation}
where $a_0$ is a parameter of the model that regulates the magnitude of the dependency from the membrane potential and $u_c$ is a cut-off threshold such that, when $I=0$ and $u > u_c$, the membrane potential grows until the emission of a spike.

Both the QIF and the EIF bring in a further level of complexity with the inclusion of non-linear dependencies that affect both the computational costs and the ease of analysis, but allow for a more precise generation of spikes \cite{Gerstner_2009}. Additionally, a hidden cost lies in the use of two extra parameters in each of them.

\begin{figure*}[h]
\centering
\subfloat[]{
    \includegraphics[width=2.5in]{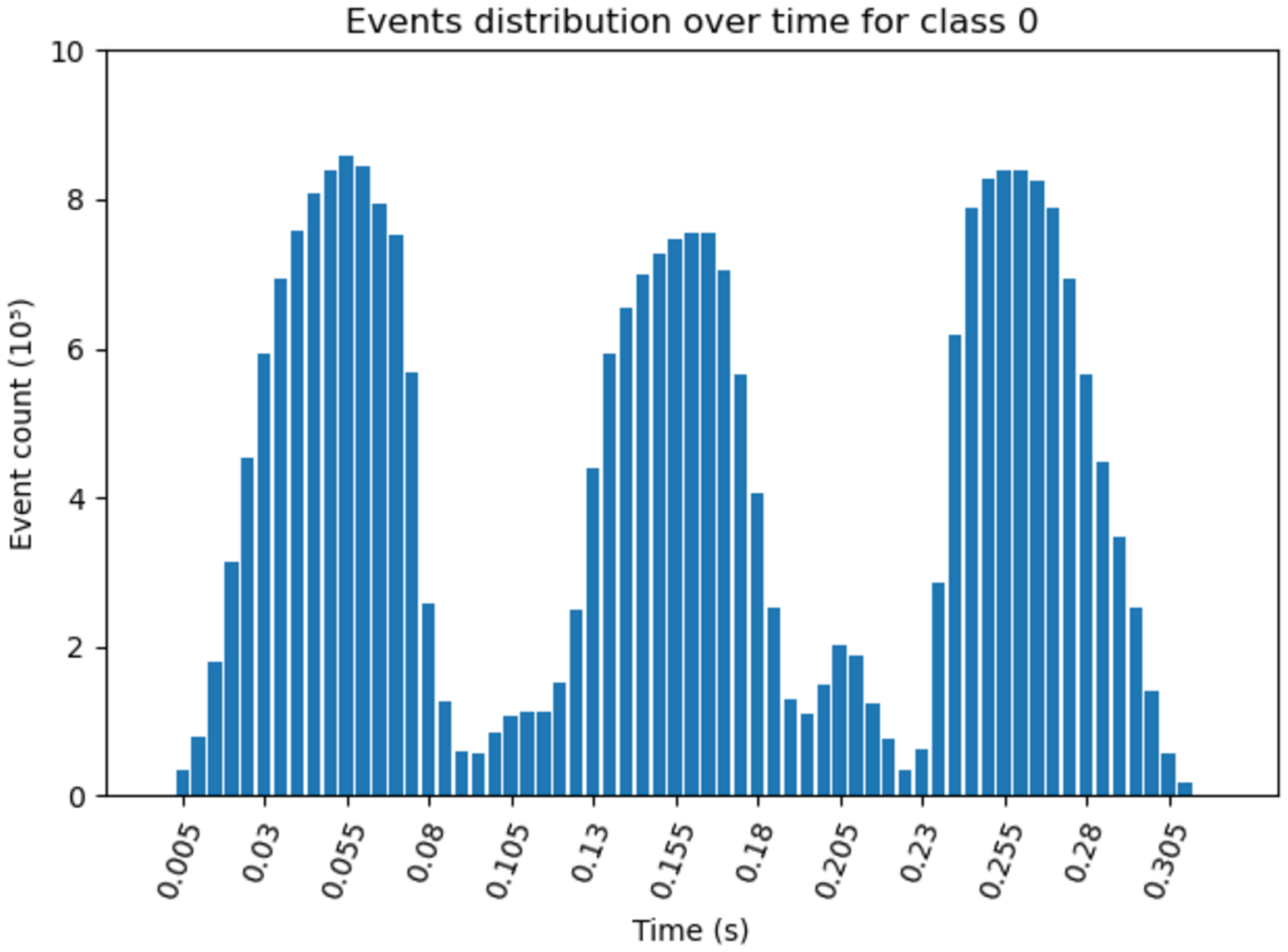}
    \label{fig_events_0}
}
\hfil
\subfloat[]{
    \includegraphics[width=2.5in]{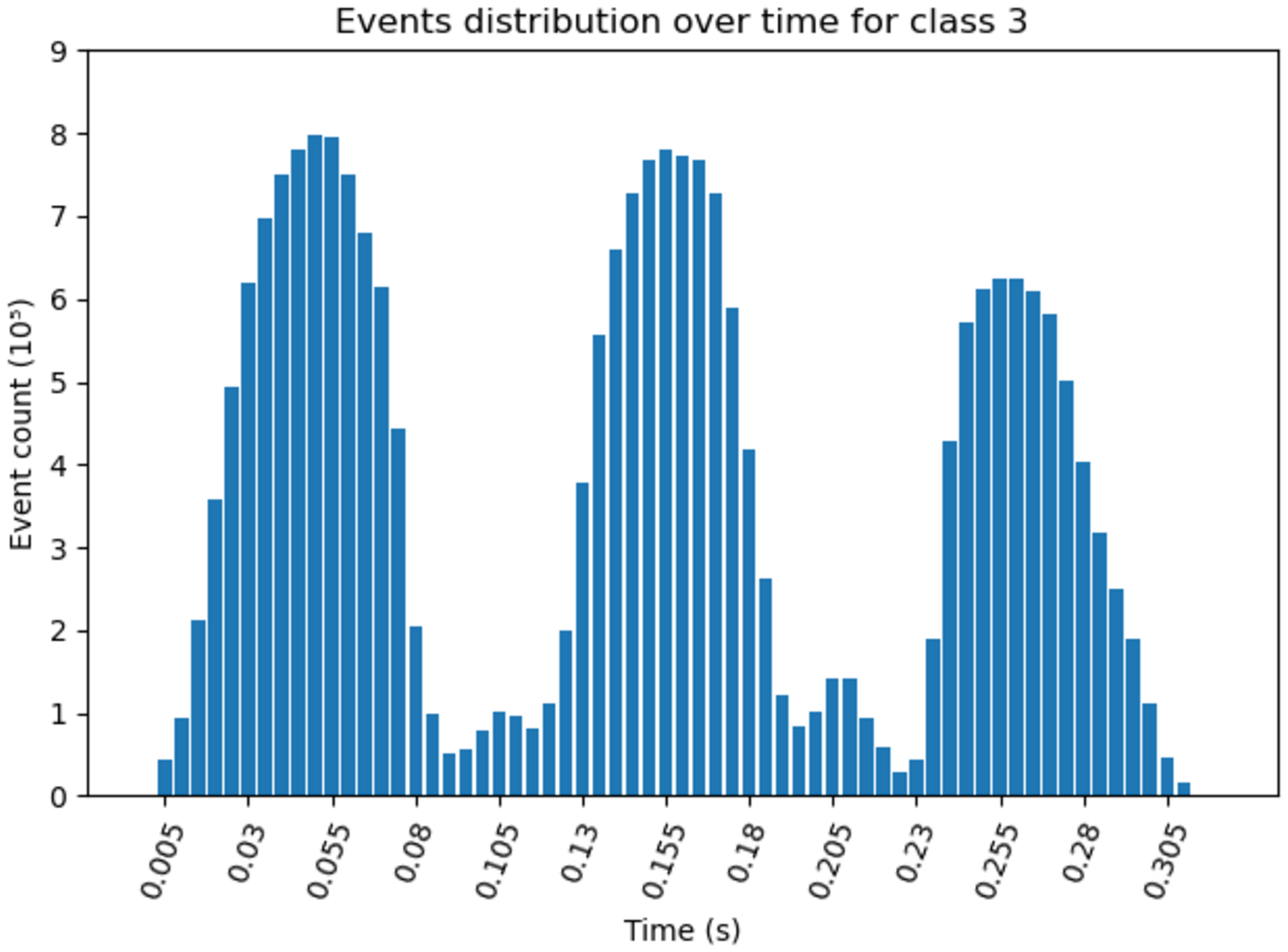}
    \label{fig_events_3}
}
\hfil
\subfloat[]{
    \includegraphics[width=2.5in]{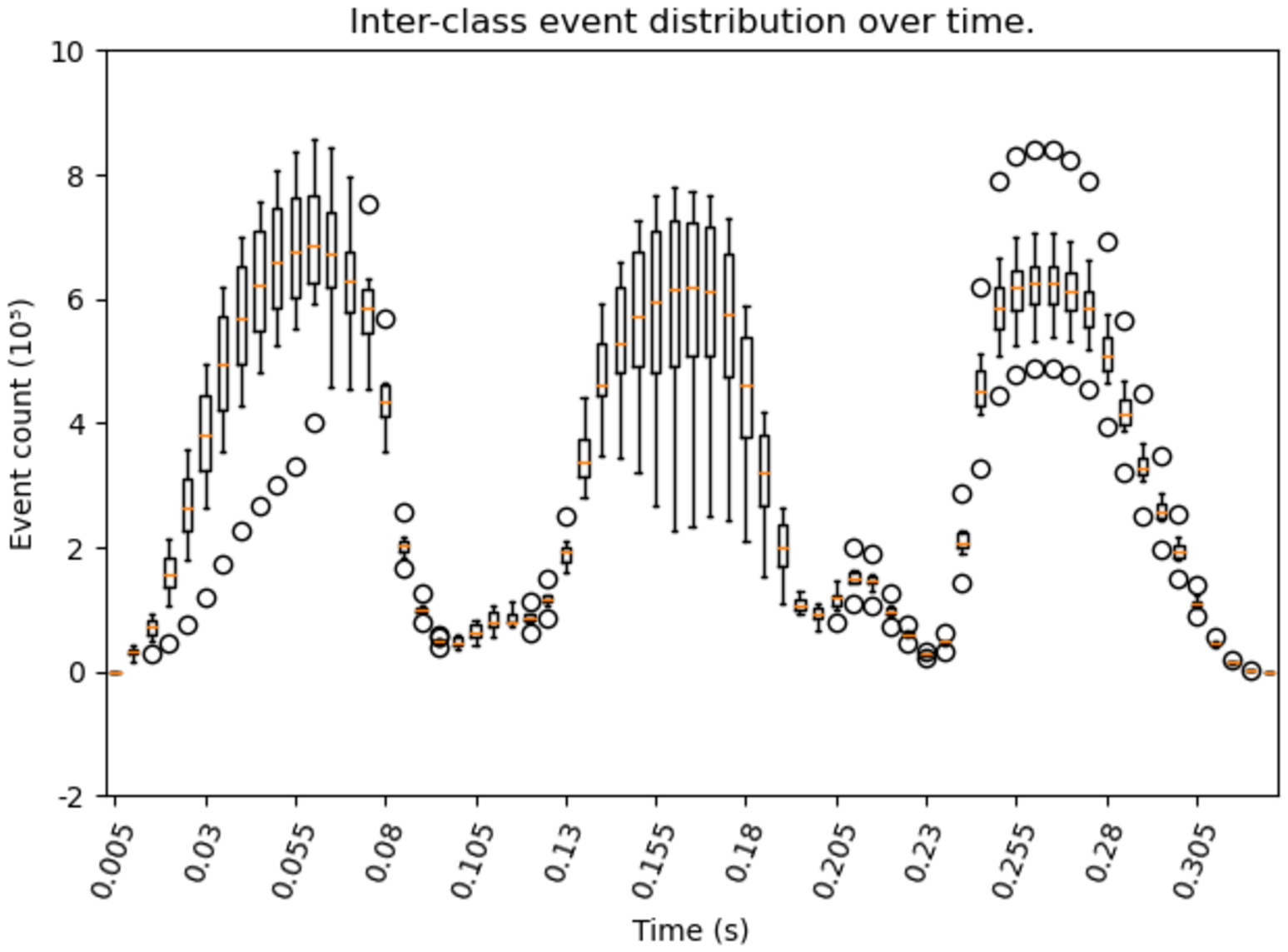}
    \label{fig_events_inter_class}
}
\caption{Visualization of the number of events over time. \figurename\  \ref{fig_events_0} and \ref{fig_events_3} for classes "0" and "3" are reported as representative. They depict the count of events for each time-step among samples of one class. \figurename\  \ref{fig_events_inter_class} reports the collective mean and variation of events throughout all the classes. As can be seen, events tend to appear always within the same time ranges for all the samples of all the classes in the dataset, thus highlighting the lack of temporal significance. Values on the y-axis are scaled by a factor of $10^5$.}
\label{fig:nmnist_events_whole}
\end{figure*}
\subsubsection{Other Multi-Variable I\&F Models}
With the inclusion of adaptation variables within the neuron model, it is possible to account for a larger number of spiking patterns and to render possible the manifestation of spike bursts, spike-adaptation responses and irregular spiking \cite{Gerstner_2009}. This comes at the cost of more differential equations in the model (one per variable) and two relevant examples are given by the Adaptive Exponential Integrate-and-Fire (AdEx) \cite{Brette2005AdaptiveExponentialIntegrateandFireModelAsanEffectiveDescriptionofNeuronalActivity} neuron model, which builds on top of the EIF, and by Izhikevich's neuron model \cite{Izhikevich_2003} which builds on top of the QIF.

A number of neuron models have been theorized in the literature, all answering to different modelling needs or considering different aspects of the observed neuronal behaviours. The ones cited above are amongst the most relevant for what concerns this study and NM computing. In fact, as reported above, the LIF model is the most widely used in the development of SNN for NM applications, but at the same time, models like the AdEx and Izhikevich's have received a lot of attention in the literature. The EIF and the QIF are on one hand the baseline of the AdEx and Izhikevich's models respectively, and, on the other hand, a slightly more complex single-variable alternative to the LIF neuron model. As such, in this study, we will focus on these single-variable models.

\section{Methods}\label{sec:methods}
We are interested in assessing the performance of different neuron models within the context of a Spiking Convolutional Neural Network (SCNN) trained with STDP. Since many factors could determine the outcome of the training, we begin by designing a simple experiment which involves the minor number of structural elements possible. This is done in order to limit the number of components that might impact the overall system performance. Therefore, we use a single-layer convolutional network in which spiking neurons are embedded right after the convolution operation on the input. 
The task set to the SCNN is a binary classification task, with the pairs of classes taken from the Neuromorphic MNIST dataset \cite{Orchard2015ConvertingStaticImageDatasetstoSpikingNeuromorphicDatasetsUsingSaccades} and the DVS Gestures dataset \cite{Amir2017ALowPowerFullyEventBasedGestureRecognitionSystem}, which contain event-based data samples. 
To develop the learning pipeline, we utilize SpykeTorch \cite{Mozafari2019SpykeTorch} as a base framework and build on top of it to include the elements required by this study, such as the diverse spiking neuron models.

\subsection{Event-based Data}
\begin{figure*}[t]
\centering
\includegraphics[width=\textwidth]{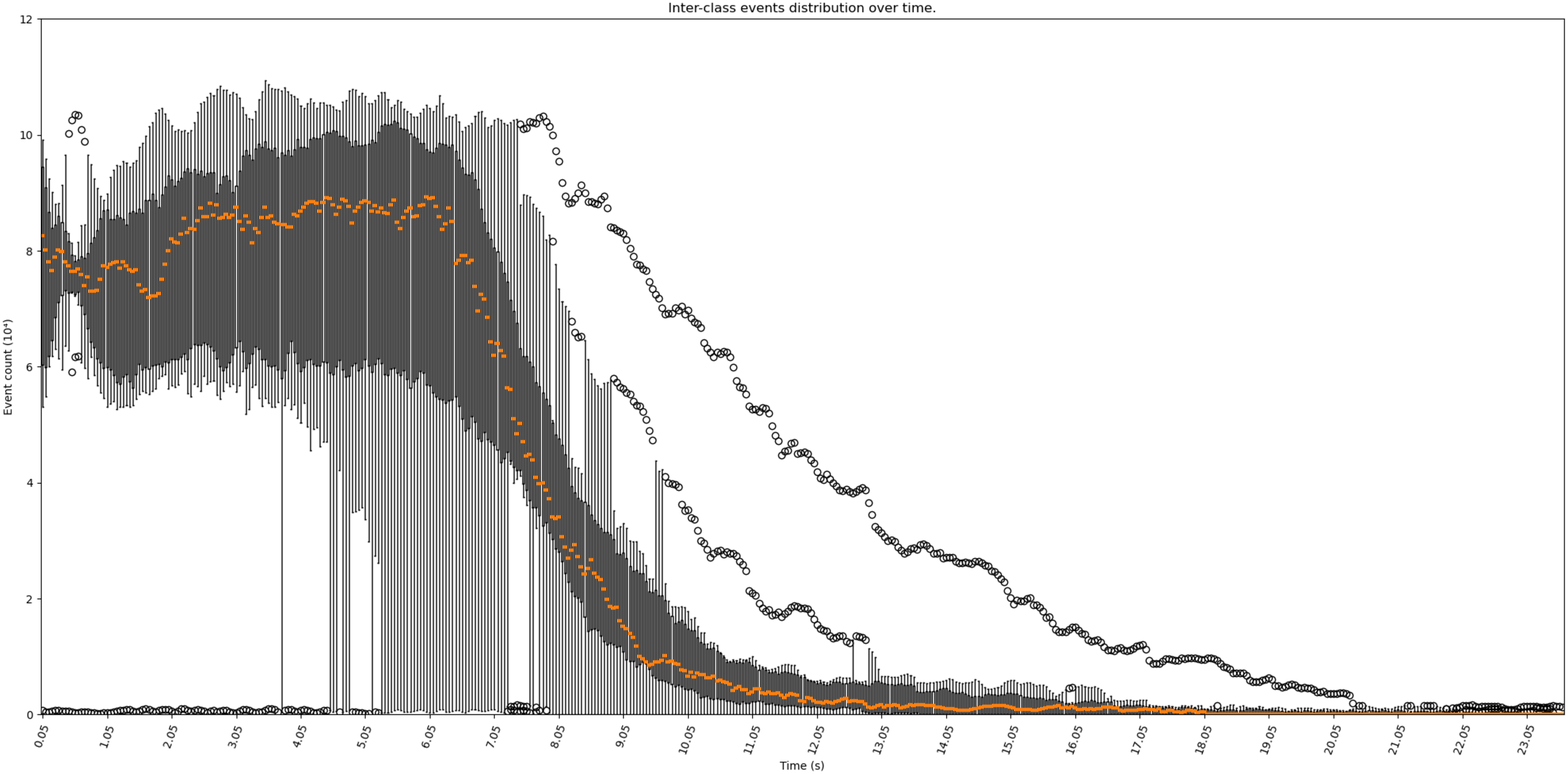}
\label{fig_events_gestures}
\caption{Visualization of the number of events over time in the DVS Gestures dataset. The events span across the whole time domain and are very dense for about the first 7 seconds. At that point, they start to decrease in number, however, the tail of the curve continues for a long time. This is due to some of the gestures lasting longer than others.}
\label{fig:inter_class_gestures_events}
\end{figure*}

To assess the performance of our simple network, we select the two natively neuromorphic datasets mentioned above. We purposely discard other non-native NM datasets as they do not possess a temporal domain, nor data is originally event-based. 
\\
\\
Data in the N-MNIST dataset is collected by recording MNIST digits shown on a screen using a moving DVS camera. Specifically, the camera makes the same 3 predefined movements for every sample, each lasting roughly 100 ms. In this way, although the dataset is built on top of a non-neuromorphic one, data samples in the dataset are natively event-based, rather than being converted from a static image. \figurename\ \ref{fig:nmnist_events_whole} reports the count of events per time for two example classes and throughout all the classes of the dataset.
By contrast, data samples in the DVS Gestures dataset (see \figurename\ \ref{fig:inter_class_gestures_events} for the inter-class distribution of events over time) are recorded using a fixed DVS camera in front of which participants move their arms according to instructions. Thus, 11 different classes of gestures are obtained, including for example arm rotation, waving or performing air guitar. The 11th class encodes "Other" random movements and is not considered in this work for simplicity.

Event data comes in the form of Address Event Representation (AER)-encoded files in which every sample is constituted by a sequence of events. Events are characterized by the specific time at which they occurred, by the location on the 2D plane and by the polarity (negative or positive light change). Similarly to \cite{Cheng2020LISNNImprovingspikingNeuralNetworksWithLateralInteractionsForRobusObjectRecognition}, to make data usable by a 2D Convolutional Neural Network (CNN) we populate a 2D image using all the events that took place between time $t$ and $t + dt$, allowing at most 1 event per (x, y) coordinates. For simplicity, we consider all events as being positive and use a batch size of 1. As a result, the network processes only 1 event map with a time resolution $dt$ belonging to only 1 sample at a time. 
Finally, we characterize each event with a value of $\frac{1}{t_s}$ in line with \cite{Gerstner_2009}. This is done in order to preserve the amount of charge that a spike carries regardless of the time-step ($t_s$) size.
No further pre-processing is applied to the data. 

\subsection{Spiking Neurons Implementation}
The phenomenological family of neuron models is the best option when developing spiking neural networks, as outlined in \sectionname\  \ref{sec:background_neurons}. We specifically concentrate on three integrate and fire neurons, namely the LIF, the QIF and the EIF. 
The LIF is the most widely used neuron that embeds a time dependency through the membrane potential leakage. The QIF and the EIF represent valid alternatives given their ability to best fit observed cortical neurons \cite{Gerstner_2009, Izhikevich2010DynamicalSystemsinNeuroscience}. Since they are single-variable models, they stand for a fairer comparison with the LIF, which is single-variable too. Indeed they all depend on the value of the membrane potential, but they all employ different types of dependencies from it. Furthermore, they are the base on which other more complex and popular neuron models are built on, respectively Izhikevich's neuron and the AdEx neuron model.

The SpykeTorch framework comes with a simple version of a I\&F neuron model.
To enable the use of these neuron models for our experiments, we expand on the framework and implement the above models by adapting the equations provided in \cite{Gerstner_2009}. State updates in the neurons are thus calculated on a per time-step basis, where each call to the neuron layer corresponds to an advancement of $t_s$ time from the previously calculated update.
Each neuron layer generates a number of neurons that reflects the size of the incoming post-synaptic currents, multiplied by the number of neuron populations that was specified at creation time. This allows for a more seamless inclusion in any point of an SNN.
When a neuron in a population emits a spike, all the other neurons belonging to the same population are inhibited and put in a refractory state to promote learning in other populations.
Each neuron model is implemented as a class inheriting from a parent Neuron class, in an object-oriented programming style. This differs from the original SpykeTorch implementation style, however, this approach was required for neurons that maintain an internal state. Besides, compatibility with the modules and neurons in SpykeTorch is maintained.
Further features and details are present in the actual implementation, however, these are not relevant to the current study and the authors point the reader to the dedicated repository page for in-depth descriptions.

\begin{figure}[!t]
    \centering
    \includegraphics[width=\columnwidth]{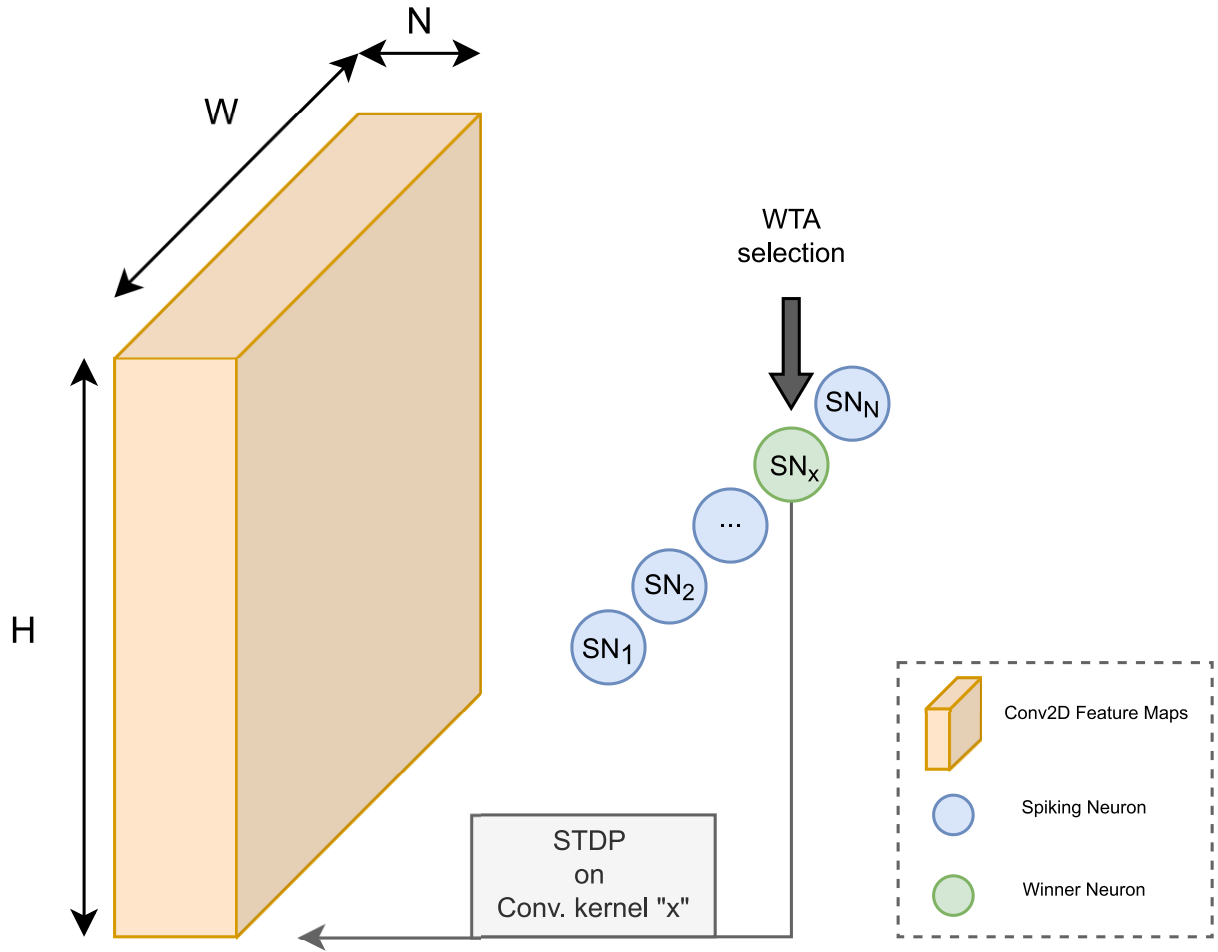}
    \caption{Diagram of the Learning Pipeline. A 2D convolution layer parses the input spike map and produces N feature maps width size WxH. Each value in each feature map is fed to a distinct neuron and the one spiking earliest is chosen as a winner by the WTA mechanism. STDP weight updates are then applied to the convolution kernel corresponding to that neuron. For ease of visualization, neuron populations are here represented as individual neurons, but each of them actually contains WxH neurons, i.e. like one feature map.}
    \label{fig:learning_pipeline}
\end{figure}
\subsection{SCNN and Learning}
We develop a simple feedforward convolutional network for our experiments, with a single convolutional layer that parses the input and connects it to the spiking neurons. \figurename\  \ref{fig:learning_pipeline} illustrates the pipeline for a better understanding. The weights of the kernel can be thought of as being synapse strengths and the resulting feature map is the input to a set of spiking neurons arranged accordingly.
In other words, the convolution represents the connectivity scheme for the spiking neurons. The use of a convolutional layer to connect inputs with the spiking neurons allows them to more easily learn spatial features.
The weights of the convolutional layer are the only parameters being learnt by the network and no pooling nor normalization is applied. 
\\
\\
We use the STDP presented and used in  \cite{masquelier2007SimplifiedSTDPRule, Kirkland2020SpikeSEGSpikingSegmentationViaSTDPSaliencyMapping, Kirkland2021UnsupervisedSpikingInstanceSegmentationOnEventDataUsingSTDP} as an unsupervised learning rule. By employing this kind of learning rule and, therefore, using it to evaluate neurons, we want to take a step closer to local feedforward learning, which is believed to be the key to exploit the potentials of neuromorphic engineering at its best \cite{Taherkhani2020AReviewofLearninginBiologicallyPlausibleSpikingNeuralNetworks, Schuman2017ASurveyofNeuromorphicComputingandNeuralNetworksinHardware}.
The mentioned STDP rule applies the following weight updates:
\\
\\
\begin{equation}
    \centering
    \Delta W_{i,j} = \begin{cases}
    A^+ \times(W_{i,j}-LB)\times(UB-W_{i,j})\quad\text{if}\quad T_j\leq T_i,\\
    A^- \times(W_{i,j}-LB)\times(UB-W_{i,j})\quad\text{if}\quad T_j > T_i,
    \end{cases}
\end{equation}

where $W_{i, j}$ is the weight of the synapse connecting neuron $j$ (pre-synaptic) to neuron $i$ (post-synaptic), $LB$ and $UB$ are a lower and an upper bound value respectively, $T_j$ is the timing of the spike emitted by neuron $j$, $T_i$ the timing of the spike emitted by neuron $i$, and $A^+$ and $A^-$ are two parameters used to scale the weight update.

It is thus a system of two parabolic equations that are applied depending on whether Long Term Potentiation (LTP) or Long Term Depression (LTD) needs to take place. One of the consequences of using this learning rule is that weight updates are self-regularized. In fact, the closer the weights get to the boundary values, the smaller the updates will be, allowing the weights to be refined more granularly as the learning proceeds. Another aspect that is important to highlight is in how this learning rule is applied. While the original theorization of Hebbian rules such as STDP states that the weight update should be proportional in value and sign on the time-difference between the post- and the pre-synaptic spikes, in a software implementation some approximations are required. Therefore, for each time-step of the execution, we apply LTP on weights connected to input locations where there has been a spike, and LTD in all the others.
In order to promote competitive and differentiated feature learning, we also employ a k-Winners Take All (WTA) (with k=1) learning paradigm. WTA allows only k neurons per time-step to be eligible for STDP updates, specifically the ones firing sooner. 
\\\\
Finally, as an homeostatic mechanism to allow neurons to keep on firing despite the changes in their synaptic weights, we re-calculate individual neurons' thresholds according to \eqref{eq:threshold_1}. This form of adaptive thresholding is often required when employing STDP, and is a common practice for SNNs \cite{Kirkland2020SpikeSEGSpikingSegmentationViaSTDPSaliencyMapping}.
\begin{equation}
\label{eq:threshold_1}
    V_{thresh} = \lambda \cdot R\cdot A\cdot\frac{t_s}{\tau_m}\cdot \overline{W}\cdot(W_k \cdot H_k)\cdot n_c, 
\end{equation}
and since $\tau_m = R\cdot C$, we can equivalently re-write:
\begin{equation}
\label{eq:threshold_2}
    V_{thresh} = \lambda \cdot A\cdot\frac{t_s}{C}\cdot \overline{W}\cdot(W_k \cdot H_k\cdot N_c), 
\end{equation}
where $A$ is the amplitude of the spike, in our case assigned to be $A = \frac{1}{t_s}$, $R$ is the resistance, $C$ is the capacity, $\tau_m$ is the membrane time constant, $\overline{W}$ is the average value of the synaptic weights, $W_k$, $H_k$ and $N_c$ are the width, height and depth (number of channels) of the synaptic kernel, and $\lambda$ is a regulation parameter that takes values in the range $[0, 1]$.
In \eqref{eq:threshold_2}, the term $A\cdot\frac{t_s}{C}\cdot \overline{W}$, can be explained as being the average effect perceived on the membrane potential as a result of a single spike, whereas the second term, $(W_k \cdot H_k)\cdot n_c$, scales this effect to the size of the synaptic kernel. Therefore,  \eqref{eq:threshold_2} calculates what would be the average post-synaptic potential perceived in the case the input was dense with spikes. The parameter $\lambda$ serves as a regulation of what percentage of this amount would be necessary to reach before emitting a spike.

\subsection{Classification}
Because we employ an unsupervised learning rule, labels are not used at any point during the learning of the weights. However, labels are needed to classify data samples, therefore we adopt a system similar to  \cite{Diehl2015UnsupervisedLearningOfDigitRecognitionUsingSTDP, Iyer2017UsupervisedLearningOfEventBasedImageRecordingsUsingSTDP} and count, for each neuron, the number of times it spiked in response to samples having  a given label. At the end of the training phase, each neuron is assigned the label for which it spiked the most during training.
\\
During the inference phase, for each data sample, a sequence of spikes is collected and weighted depending on the order they arrived. These are then summed and the label corresponding to the highest value is selected as the classification outcome.

\subsection{Hyper-parameter Optimization}\label{sec:optimization}
Defining a good set of parameters for a machine learning system is a non-trivial task. This often requires a lot of expertise and hand tuning and is very error-prone. In order to reduce the possibility of selecting sub-optimal parameters for neurons which would result in poor performance, we thus make use of an optimization system to find reasonably good parameters for our experiments.
To this extent, we employ the BOHB optimization \cite{Falkner2018BOHBRA} using the HpBandSter library. This technique combines Bayesian Optimization (BO) and Hyperband (HB), a resource allocation and early stopping strategy. 
As a result, we can draw more robust conclusions about the performance of the neurons.

\section{Results}
We train the SCNN outlined in \sectionname\  \ref{sec:methods} using a subset of the N-MNIST and the DVS Gestures datasets. Indeed, in order to not present the network with too difficult a job, we define a binary classification task between 2 classes from the datasets. More specifically, we randomly select 4 distinct couples of classes from each dataset and define, for each, a separate binary classification task. In this way, we aim to obtain more generalizable results and to reduce the dependency of the results on a particular coupling.
Furthermore, the order in which data is presented to the SCNN might be influential in a system employing STDP as a learning rule. This is due to the fact that STDP rewards and builds on the inputs that are presented earlier. Therefore, to ensure independence from this behaviour of STDP, we repeat every experiment a total of 11 times per task.
\subsection{Same Hyper-Parameters Training}
We first conduct our experiments using hand-tuned hyper-parameters. These were found by a trial-and-error practice and represent a set of parameters that enabled learning for the task at hand. This means that neurons using these parameters were able to emit spikes and to have the weights adjusted in a way that enabled the learning of representations of the inputs.
Where possible, we adopt the same hyper-parameters for all the neurons in all the experiments on each dataset. Since the QIF and the EIF models introduce 2 different hyper-parameters each, each of them undergoes a further hand-tuning.
Results of the training sessions are shown in Table \ref{tab:same_params_nmnist} for the N-MNIST-based tasks and in Table \ref{tab:same_params_gestures} for the DVS Gesture-based tasks. Here, for every task and neuron model, are reported the average and best test accuracies achieved, calculated according to (\ref{eq:accuracy_eq}):
\begin{equation}
    \label{eq:accuracy_eq}
    accuracy = \frac{TP + TN}{TP + TN + FT + FN},
\end{equation}
where TP stands for true positive, TN for true negative, FT for false true and FN for false negative.
On each column (task), the best average score and the absolute best score are highlighted in bold.

\begin{table}[t!]
\caption{Table of results on the N-MNIST dataset. In each cell, the mean accuracy $\pm$ the standard deviation values are followed by the best accuracy found (after the comma). Values are rounded up to the closest second decimal value.}
\centering
\resizebox{\linewidth}{!}{

    \begin{tabular}{@{}ccccc@{}}
    \tr
    Neuron Model & 0 vs 1                                                       & 2 vs 9                                                      & 3 vs 7                                                       & 4 vs 8                                                       \\ \midrule\midrule
    LIF          & 
    \begin{tabular}[c]{@{}c@{}}0.77$\pm$0.11, \\ 0.93\end{tabular} & \textbf{\begin{tabular}[c]{@{}c@{}}0.74$\pm$0.06,\\ 0.79\end{tabular}} & \textbf{\begin{tabular}[c]{@{}c@{}}0.73$\pm$0.04,\\ 0.78\end{tabular}}  & \begin{tabular}[c]{@{}c@{}}\bf0.57$\pm$0.02,\\ 0.59\end{tabular}  \\ \mr
    EIF          & 
    \textbf{\begin{tabular}[c]{@{}c@{}}0.80$\pm$0.12, \\ 0.94\end{tabular}} & \begin{tabular}[c]{@{}c@{}}0.64$\pm$0.07,\\ 0.71\end{tabular} & \begin{tabular}[c]{@{}c@{}}0.65$\pm$0.04,\\ 0.71\end{tabular}  & \begin{tabular}[c]{@{}c@{}}0.55$\pm$0.04,\\ \bf0.61\end{tabular}  \\ \mr
    QIF          & 
    \begin{tabular}[c]{@{}c@{}}0.57$\pm$0.03, \\ 0.61\end{tabular} & \begin{tabular}[c]{@{}c@{}}0.54$\pm$0.03,\\ 0.58\end{tabular} & \begin{tabular}[c]{@{}c@{}}0.51$\pm$0.02, \\ 0.55\end{tabular} & \begin{tabular}[c]{@{}c@{}}0.52$\pm$0.02,\\ 0.55\end{tabular} \\ \br
    \end{tabular}
    }
    \label{tab:same_params_nmnist}
\end{table}
\begin{table}[t!]
\caption{Table of results on the DVS Gestures dataset. In each cell, the mean accuracy $\pm$ the standard deviation values are followed by the best accuracy found (after the comma). Values are rounded up to the closest second decimal value. In the table, HC stands for Hand Clapping, RHW for Right Hand Wave, RACW for Right Arm Clockwise, AG for Air Guitar, RACCW for Right Arm Counter Clockwise, AR for Arm Roll, LACW for Left Arm Clockwise and AD for Air Drums.}
\centering
\resizebox{\linewidth}{!}{
    \begin{tabular}{@{}ccccc@{}}
    \tr
    Neuron Model & HC vs RHW                                                    & RACW vs AG                                                   & RHCW vs AR                                                  & LACW vs AD                                                 \\ \midrule\midrule
    LIF          & 
    \begin{tabular}[c]{@{}c@{}}0.53$\pm$0.06,\\ 0.60\end{tabular} & \begin{tabular}[c]{@{}c@{}}0.50$\pm$0.05,\\ 0.56\end{tabular} & \begin{tabular}[c]{@{}c@{}}0.54$\pm$0.13,\\ 0.66\end{tabular} & \begin{tabular}[c]{@{}c@{}}0.52$\pm$0.09,\\ 0.69\end{tabular} \\ \mr
    EIF          & 
    \textbf{\begin{tabular}[c]{@{}c@{}}0.58$\pm$0.12,\\ 0.77\end{tabular}} & \textbf{\begin{tabular}[c]{@{}c@{}}0.50$\pm$0.04,\\ 0.58\end{tabular}} & \begin{tabular}[c]{@{}c@{}}0.53$\pm$0.11,\\ 0.66\end{tabular} & \begin{tabular}[c]{@{}c@{}}0.46$\pm$0.05,\\ 0.52\end{tabular} \\ \mr
    QIF          & 
    \begin{tabular}[c]{@{}c@{}}0.57$\pm$0.10,\\ 0.71\end{tabular} & \begin{tabular}[c]{@{}c@{}}0.48$\pm$0.04,\\ 0.52\end{tabular} & \textbf{\begin{tabular}[c]{@{}c@{}}0.62$\pm$0.06,\\ 0.67\end{tabular}} & \textbf{\begin{tabular}[c]{@{}c@{}}0.53$\pm$0.09,\\ 0.75\end{tabular}} \\ \br
    \end{tabular}
    }
    \label{tab:same_params_gestures}
\end{table}

By examining the results on the N-MNIST-based tasks in Table \ref{tab:same_params_nmnist}, the LIF neuron model is found to perform better than the other two on average. Indeed, the EIF has higher average accuracy only on the 0 vs 1 task, whereas the QIF model fails to achieve accuracy levels high enough to match any of the other two counterparts.

Considering the results reported in Table \ref{tab:same_params_gestures}, the situation differs slightly. The performance of both the LIF and the EIF neuron models, on average, decreases drastically, whereas the QIF maintains similar levels of accuracy as in the N-MNIST case. Nevertheless, both the EIF and QIF demonstrate superior classification abilities throughout and their top accuracy levels often surpass those of the LIF model.
These trends in the accuracy levels highlight two main aspects. Firstly, the complexity of the N-MNIST data is lower than that of the DVS Gestures dataset. Indeed, in both cases, the very same neural network architecture and neuron models were employed, yet the accuracy in the DVS Gestures is on average considerably lower, hence highlighting the greater difficulty of the task. Data samples in the DVS Gestures dataset arguably have richer visual and temporal features which render the learning more difficult when compared to the N-MNIST. Secondly, the richer temporal diversity of the features might be better represented by means of neurons with richer voltage dynamics, such as the QIF and EIF. As reported by the experiments, in fact, these two are steadily better than the LIF models and even though in some instances one performs more poorly, the other still attains higher accuracy, possibly as a result of a better affinity to the temporal dynamics found in that particular task.
\subsection{Optimized Hyper-Parameters Training}
\label{sec:optim_results}
As a second set of experiments, we employ the optimization system outlined in \sectionname\  \ref{sec:optimization} to obtain a set of hyper-parameters that is heuristically optimal for a specific scenario. The optimization is carried out on the "0 vs 1" and on the "HC vs RHW" tasks for each neuron model individually. We allow a total of 24 optimization iterations for each neuron and task, to not favor any experiment over the others. Results are reported in Table \ref{tab:optim_params_both}. Once again, after obtaining the optimized hyper-parameters, we train and evaluate each model a total of 11 times to increase the robustness of the results.
\begin{table}[t!]
\caption{Table of results using optimized hyper-parameters. In each cell, the mean accuracy $\pm$ the standard deviation values are followed by the best accuracy found (after the comma). Values are rounded up to the closest second decimal value. Optimization and evaluation is performed on one representative task per dataset only.}
\centering
    \begin{tabular}{@{}lcc@{}}
    \tr
    Neuron Model      & 0 vs 1                                                          & HC vs RHW                                                                                                                   \\ \midrule\midrule
    LIF      & 
    \begin{tabular}[c]{@{}c@{}}\bf0.95$\pm$0.02,\\ 0.982\end{tabular} &
    \begin{tabular}[c]{@{}c@{}}0.53$\pm$0.05,\\ 0.625\end{tabular}  \\
    \mr
    EIF & 
    \begin{tabular}[c]{@{}c@{}}0.74$\pm$0.15,\\ 0.93\end{tabular} &
     \textbf{\begin{tabular}[c]{@{}c@{}}0.57$\pm$0.11,\\ 0.67\end{tabular}} \\
     \mr
    QIF &  
    \begin{tabular}[c]{@{}c@{}}0.90$\pm$0.05,\\ \bf0.985\end{tabular} & \begin{tabular}[c]{@{}c@{}}0.55$\pm$0.05,\\ 0.625\end{tabular} \\ 
    \br
    \end{tabular}

    \label{tab:optim_params_both}
\end{table}
\\\\
Since the optimization is task-specific, we only evaluate our models on the two representative tasks they were optimized on. In the case of the "0 vs 1" task based on the N-MNIST, overall, the accuracy levels drastically increase. The LIF model accuracy grows by nearly 20 percentage points on average; however, the most striking increase is the accuracy of the QIF model, which gains 33 percentage points on average and 37.5 in the best case. This not only highlights the sensitivity of neuron models to their hyper-parameters, but also confirms that neurons with more complex dynamics can perform just as well as simpler ones. 
\\
In the case of the "HC vs RHW" task, instead, we see a slightly different trend. In the first place, the results are surprisingly worse than those obtained through hand-picked parameters. We hypothesize that this is because the optimization system required more iterations to find a good set of hyper-parameters. As stated above, we allowed the same number of optimization iterations as in the case of the N-MNIST dataset. However, given the higher complexity of the features present in the DVS Gestures dataset, it might have been necessary to allow more. Secondly, although still struggling to achieve higher accuracy levels, the SNNs employing QIF and EIF averagely outperform those with the LIF. This confirms the results obtained using the same hyper-parameters and strengthen the hypothesis that richer dynamics can be beneficial when employed on data with a richer set of temporal features. 
\\\\
Another point worth considering is the variability of the results obtained. Spanning from the N-MNIST-based tasks to the Gestures-based ones, the different neuron models demonstrate accuracy levels with a standard deviation of up to 15 percentage points. We hypothesize that this effect is caused by the order in which data is presented to the system in relation to the STDP learning rule which, as reported at the beginning of this section, is sensitive to such order. Apart from this, we further observe that the EIF model has higher fluctuations on average, thus demonstrating more sensitivity to this effect.
\section{Discussion}
\subsection{Implications of Using Different Neuron Models}
The usage of spiking neuron models has some inherent implications on the machine learning pipeline from the implementation and the theoretical points of view.
\\
Concerning the implementation, spiking neuron come with a whole set of hyper-parameters to tune. Considering the LIF, the simplest version requires a single parameter (the time costant or a leakage term), but other implementations might include up to 5 different parameters, such as the refractory period or the time-step size. If we switch to the QIF or the EIF, there are at least two new and non-optional parameters to consider (see \ref{sec:background_neurons}).
\\
Determining a good set of hyper-parameters is a non-trivial task \cite{Gerstner2009HowGoodAreNeuronModels}. Although in the NM field a lot of inspiration is taken from the human brain, it is not possible to simply assume that the same parameters that work in such a complex system would still be applicable in a simplification such as an SNN. We hence need to tweak parameters for our need or, alternatively, to define a parameter optimization strategy that does that heuristically in an automated way. However, also the latter solution often requires to make guesses about the domain in which parameters can vary and it requires a long time to compute. Furthermore, hyper-parameters can be correlated in some way, thus making both the hand-tuning and the automated optimization process more difficult.
As a result, from an implementation point of view, using neuron models that require more hyper-parameters can significantly increase the usage complexity.
\\
\begin{figure*}[!t]
    \centering
    \subfloat[]{
    \includegraphics[width=3in]{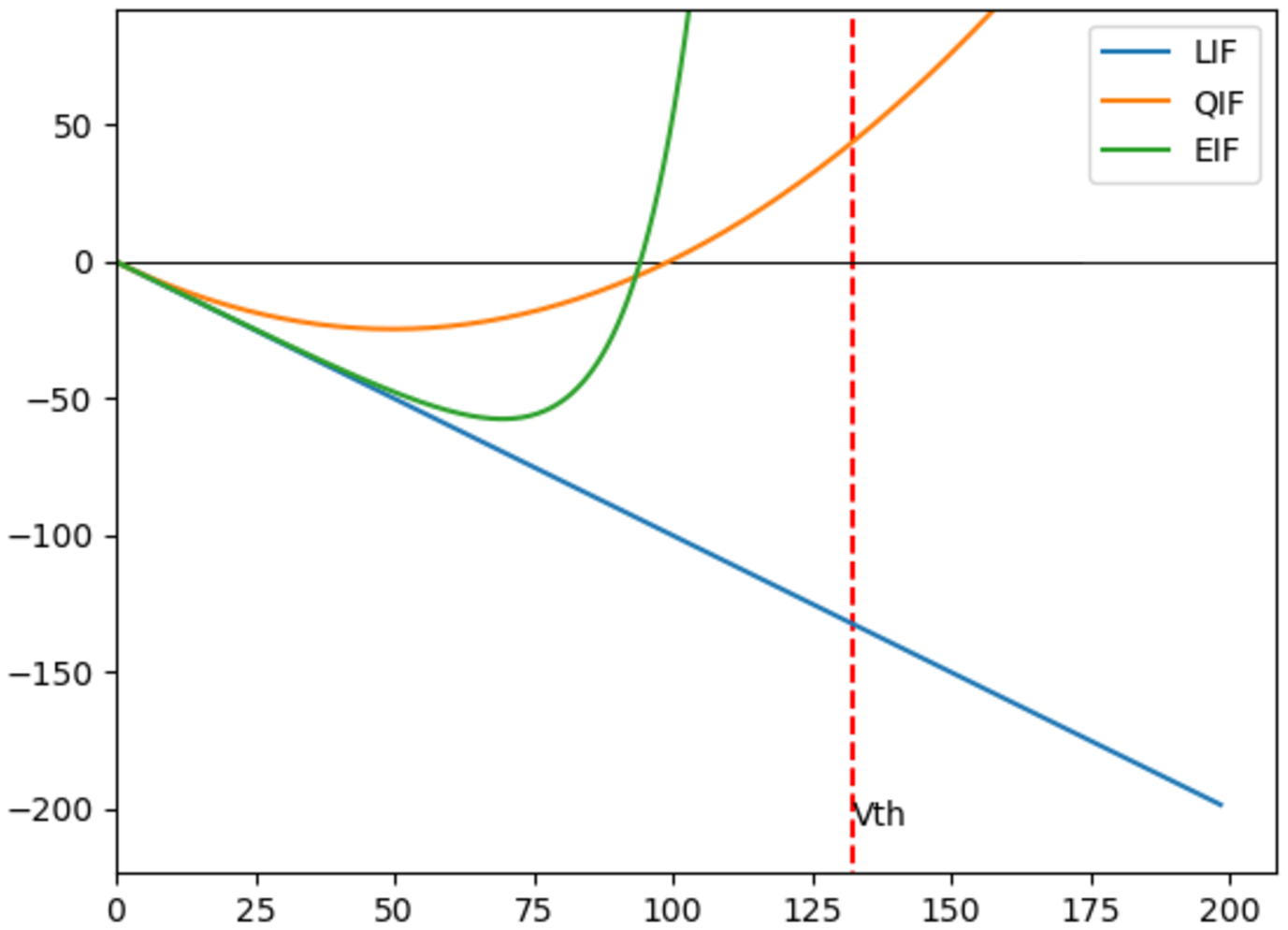}
    \label{fig:LIF_QIF_EIF_odes}
    }
    \hfil
    \subfloat[]{
        \includegraphics[width=3in]{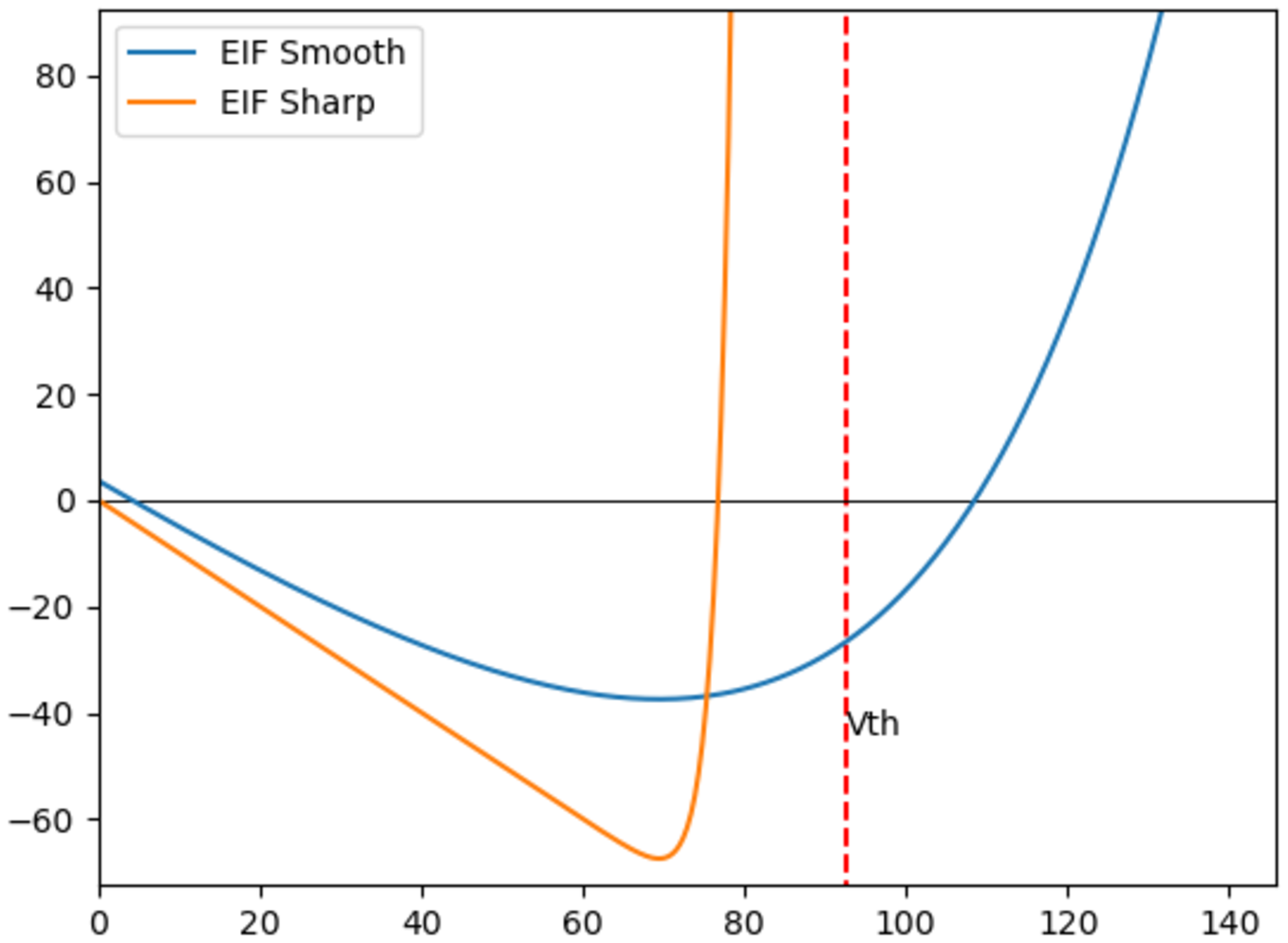}
        \label{fig:EIF_variants}
    }
    
    \caption{Example of membrane potential dynamics of the spiking neuron models. In both figures, the y-axis represents the variation of the membrane potential $du$, while the x-axis represents the value of the membrane potential itself $u$. \figurename\  \ref{fig:LIF_QIF_EIF_odes} compares the three spiking neurons, whereas \figurename\  \ref{fig:EIF_variants} is an example of how varying the sharpness parameter $\Delta_T$ can affect the dynamics of the EIF.}
    \label{fig:neurons_odes}
\end{figure*} 

\begin{figure}[!t]
    \centering
    \subfloat[]{
    \includegraphics[width=\columnwidth]{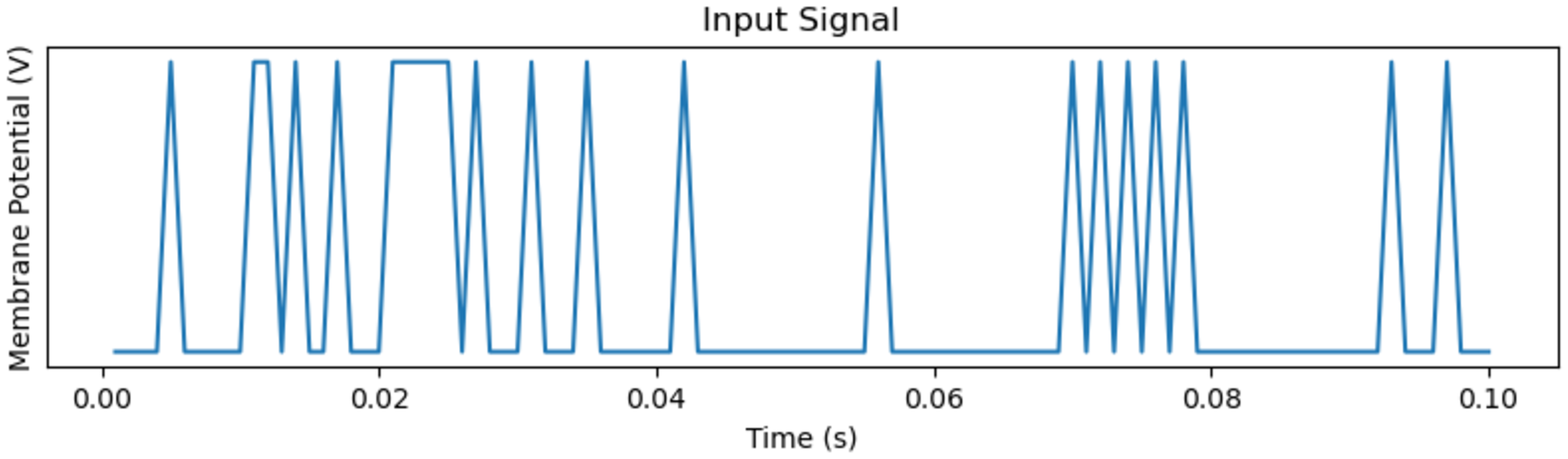}
    \label{fig:fast_slow_lif_input}
    }\hfil
    \subfloat[]{
    \includegraphics[width=\columnwidth]{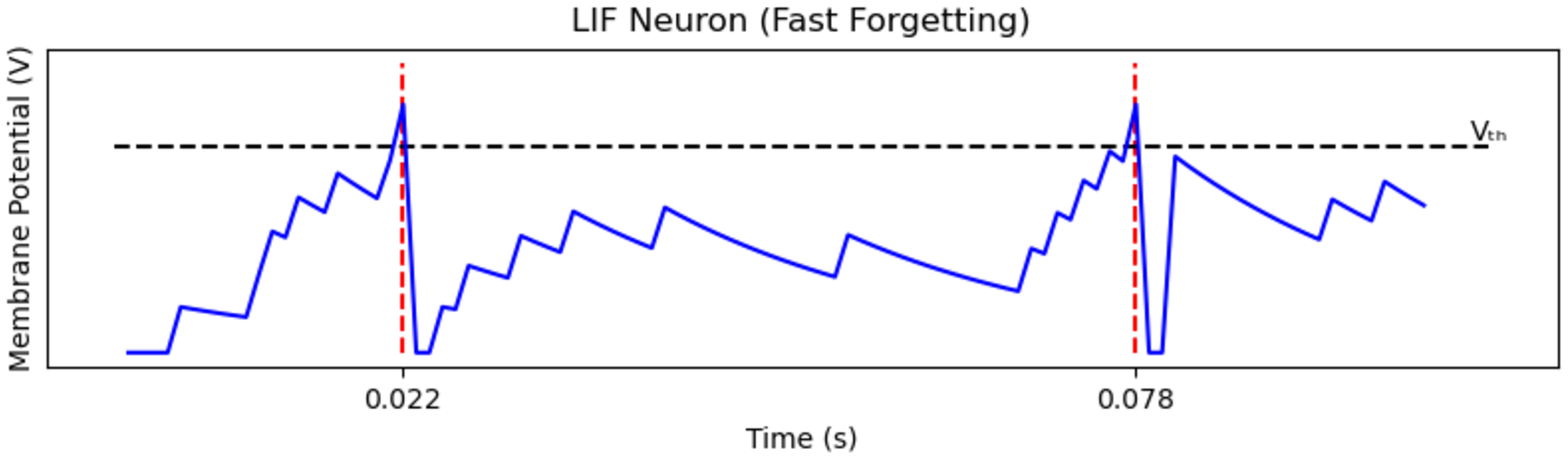}
    \label{fig:fast_slow_lif_fastforgetting}
    }\hfil
    \subfloat[]{
    \includegraphics[width=\columnwidth]{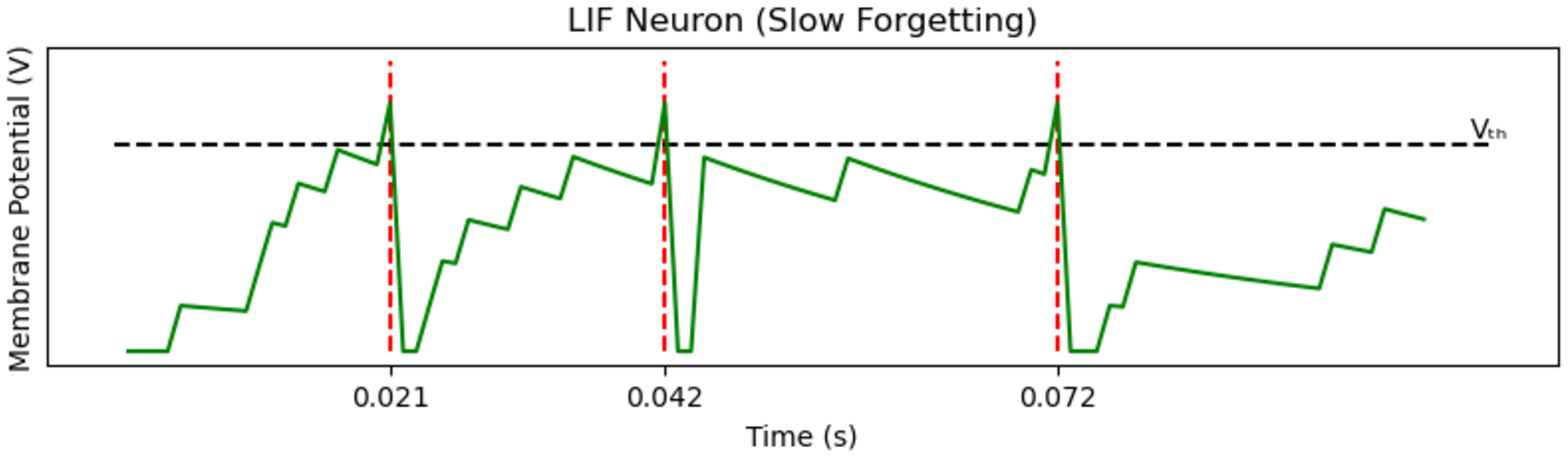}
    \label{fig:fast_slow_lif_slowforgetting}
    }
    \caption{Example of Changing Hyper-Parameters in a LIF Neuron. The ``Fast forgetting'' neuron (smaller time constant $\tau_m$) (\figurename\ \ref{fig:fast_slow_lif_fastforgetting}) can only spike twice in response to the input (\figurename\ \ref{fig:fast_slow_lif_input}). The ``Slow forgetting'' one (\figurename\ \ref{fig:fast_slow_lif_slowforgetting}) can fire three times and maintains a higher membrane potential throughout. Note that the ``Slow forgetting'' neuron also fires earlier, i.e. it requires less (close) spikes to reach the threshold. Both the neurons have a refractory period of 2 ms.}
    \label{fig:fast_slow_lif}
\end{figure}
From a theoretical point of view, using different neurons or varying the parameters means to open to different non-linear dynamics and excitability patterns. \figurename\  \ref{fig:neurons_odes} and \figurename\  \ref{fig:fast_slow_lif} provide a visual understanding of these differences.
In the LIF, the membrane potential updates depend linearly on the previous state of the membrane potential itself. The EIF manifests a similar relationship up to certain values of membrane potential ($\Theta_{rh}$), after which the relationship assumes a more non-linear (exponential) aspect. The QIF loses any linear relationship in favor of a quadratic one. 
These dynamics play a role on the excitability (regions) of a neuron \cite{Gerstner_2009}. For instance, an EIF with a smooth exponential term (blue line in \figurename\  \ref{fig:EIF_variants}) will receive more mitigated updates throughout (slow-forgetting neuron), whereas an EIF with a sharp exponential term (orange line) will receive more negative updates up to the cutoff threshold $\Theta_{rh}$ and then very positive ($+\infty$) ones, thus immediately reaching the firing threshold $V_{th}$. Hence, the second example would be a fast-forgetting neuron, but the cut-off threshold will act as an early firing threshold, as any subsequent update would bring the membrane potential up and above the actual firing threshold.
A similar example is reported in \figurename\  \ref{fig:fast_slow_lif}, where a change in the time-constant $\tau_m$ makes a LIF neuron forget faster or slower. This in turn has effects on the excitability of the neuron and its firing pattern.
\\
\\
The different firing abilities discussed above need to be considered within the context where the neurons are placed. If we consider the case of a homogeneous SCNN that is trained on a dataset in which the temporal distribution of events is very similar for every sample, it might be pointless to consider having a wide range of excitability patters as more complex neurons have. Indeed, the increased amount of parameters would make it very difficult to find the right excitability that works well with that data. At the same time, they would likely come at a higher computational and power cost. Conversely, if the dataset is very diverse in terms of temporal distribution of events, it would arguably prove useful to have a broad range of excitability patterns to choose from. Therefore, an heterogeneous network of spiking neurons would likely be able to learn better or simply more features. In this context, employing more complex neurons with variable parameters can be significantly beneficial.

\subsection{Temporal Features and Neuron Performance}
In our experiments, we used a very simple homogeneous SCNN to perform a simple classification task on a simple subset of the N-MNIST and DVS Gesture datasets. The N-MNIST dataset, although natively event-based, is not a naturally dynamic dataset. The original data, the MNIST handwritten digits, are static images that do not contain temporal dynamical features. As such, the temporal features that are instead present in the N-MNIST are crafted. Furthermore, these dynamics are obtained by moving a DVS camera using the same sequence of movements with the same timing. \figurename\  \ref{fig_events_inter_class} shows that, as a result, the distribution of events throughout each sample present in the dataset is roughly the same.
This means that the temporal features are not different from one another and are hence not discriminative of different classes of samples.
Indeed, as discussed in the previous paragraph, using a homogeneous SCNN was enough to achieve reasonably high accuracy levels, despite the lack of diversity in the dynamics of the embedded neurons.

Concerning the performance in such a homogeneous settings, in our consideration of three single-variable neuron models we found that all of them have the ability to perform well. The difference however is in the cost of using one neuron rather than the other. From our experiments, we found that when hand-tuning parameters, the LIF neuron achieved averagely high accuracy levels, with EIF neuron being better at times. Using a set of optimized hyper-parameters, we observed a considerable improvement in the overall classification accuracy with the QIF achieving a 98.5\% accuracy in the best case, thus surpassing its counterparts. The very same QIF model performed rather poorly when using non-optimized parameters.
As mentioned in \sectionname\  \ref{sec:optim_results}, this highlights the fact that, despite the data displaying simple spatio-temporal features, more complex neurons are still able to perform well.
The cost for achieving such results can, however, become rather high. 
\\\\
When employing Gestures data, the situations is slightly different. In this case, as depicted in \figurename\  \ref{fig:inter_class_gestures_events}, there is no recurring distribution of events across different classes. Instead, events are distributed throughout the whole time domain. Each class of gestures has a distribution of events that varies with respect to the others, even more because of the fact that different actions require a different time to be executed. Thus, the temporal features in this dataset are more important and diverse.
As a matter of fact, this is also shown by the results obtained using the same network as in the previous case. Here, although the performance gain is still modest, the EIF and QIF neurons steadily attain better classification accuracies than the LIF model. Since we used the same setting for all the experiments, this is likely traceable to the aforementioned differences in temporal features, which are now more diverse and complex than those in the N-MNIST dataset.
\\

\subsection{Temporal Features and Depth of the Network}
The matter of temporal variety in the features being better represented by more complex dynamics opens up further questions as to their use in SNNs. Indeed, if we consider a hierarchical NN, the deepest layers normally learn more abstract representations, whereas the early layers typically learn to distinguish simple patterns, such as edges or corners \cite{Goodfellow-et-al-2016}.
This is easily conceivable when thinking about spatial features. For example, when a set of lines is recognized in the early layers of a CNN, these could later be understood to be a square, and further down the network as a house. Although it can be more difficult to imagine, when we include the time dimension, similar scenarios can arise where features relate temporally other than spatially. It thus seems straightforward that the temporal relationships might vary in complexity in different stages of the network depending on the task at hand.\\
We have proved that the use of more complex neuron models improves the performance on more complex tasks at the level of one layer. When combining several of these layers in a hierarchical network more uses of their non-linear dynamics could arise, as they would combine several spatio-temporal features built up in previous stages to understand compound featural patterns.





\section{Conclusion}
In this work, we have considered a simple unsupervised SCNN and analyzed the effect on the overall performance of changing the underlying neuron models. Because they were not previously present in the SpykeTorch framework, we implement the spiking neurons and make the code available. We firstly draw a set of 4 binary classification tasks using 4 couples of classes from the N-MNIST dataset on which we repeatedly train and evaluate our SCNN. Experimental results on these tasks show that all three neuron models can achieve top-level accuracies, albeit the more complex ones require more fine-tuning. On a second instance we consider the DVS Gestures dataset, which exposes a richer set of features from both the visual and temporal points of view. In this case, the EIF and QIF steadily outperform the LIF on all 4 tasks drawn from this dataset.
Collectively, our results show that accurately selecting the neuron model employed in an NM pipeline improves its performance, and that this selection should be driven by considering the complexity of the spatio-temporal features that the layer in the network will have to understand. Furthermore, it highlights that further research aimed at unveiling the role of the dynamics of neuron models in deep hierarchical learning would be highly beneficial to close the gap between conventional DL approaches and SNNs. Other future studies could consist of analysing further relationships between the neuron models and other components of the learning pipeline, such as the neural network architecture, and the learning rule.



        



%



\section*{Acknowledgements}
This work was supported by the US Air Force Office of Scientific Research under Grant for project FA8655-20-1-
7037. The contents were approved for public release under case AFRL-2022-2603 and they represent the views of only the
authors and does not represent any views or positions of the
Air Force Research Laboratory, US Department of Defense,
or US Government.
The authors declare that there is no conflict of interest.

  \newpage



%

\section*{References}
\bibliographystyle{unsrt}
\bibliography{./bibliography}

\end{document}